\documentclass{article}

\usepackage{arxiv}

\usepackage[utf8]{inputenc} 
\usepackage[T1]{fontenc}    
\usepackage{hyperref}       
\usepackage{url}            
\usepackage{booktabs}       
\usepackage{amsfonts}       
\usepackage{nicefrac}       
\usepackage{microtype}      
\usepackage{lipsum}
\usepackage{graphicx}
\usepackage{amsmath}
\graphicspath{ {./images/} }

\title{Group Relative Policy Optimization for Image Captioning}

\author{
 Xu Liang \\
  School of Software Engineering\\
  Xi’an Jiaotong University\\
  Xi’an, China \\
  \texttt{liangxu@stu.xjtu.edu.cn} \\
}
\date{}

\begin{document}
\maketitle

\begin{abstract}
Image captioning tasks usually use two-stage training to complete model optimization. The first stage uses cross-entropy as the loss function for optimization, and the second stage uses self-critical sequence training (SCST) for reinforcement learning optimization. However, the SCST algorithm has certain defects. SCST relies only on a single greedy decoding result as a baseline. If the model itself is not stable enough, the greedy decoding result may be relatively worst, which will lead to a high variance of advantage estimation, further leading to unstable policy updates. In addition, SCST only compares one sampling result with the greedy decoding result, and the generation diversity is limited, which may fall into a local optimum. In this paper, we propose using the latest Group Relative Policy Optimization (GRPO) reinforcement learning algorithm as an optimization solution for the second stage. GRPO generates multiple candidate captions for the input image and then continuously optimizes the model through intragroup comparison. By constraining the amplitude of policy updates and KL divergence, the stability of the model during training is greatly guaranteed. In addition, compared to SCST, which only samples one answer, GRPO samples and generates multiple answers. Multiple candidate answers in the group cover a wider solution space. Combined with KL divergence constraints, GRPO can improve diversity while ensuring model stability. The code for this article is available at \url{https://github.com/liangxu-one/ms-models/tree/image_caption_grpo/research/arxiv_papers/Image_Caption_GRPO}.
\end{abstract}


\section{Introduction}
Image captioning is a task that aims to describe the content of an image by generating a natural language description. It is a multi-modal task that combines computer vision and natural language processing. Inspired by machine translation\cite{zhang2015deep}, most image captioning methods use an encoder-decoder architecture with an attention mechanism. The encoder encodes the image into a fixed-length feature vector, while the decoder generates captions word by word in an autoregressive method. Early methods mainly use pre-trained CNN encoders such as ResNet\cite{he2016deep} to extract features from images, and then use LSTM\cite{hochreiter1997long} as a decoder to decode word by word. However, the LSTM computation process is step by step. This makes it difficult for LSTM to perform efficient parallel computation during training and inference. With the great progress made by Transformer\cite{vaswani2017attention} in the field of NLP, researchers began to use Transformer as a decoder in image captioning. The Transformer model is based on the self-attention mechanism and can be trained in parallel, so using Transformer as a decoder can greatly improve the training speed.

However, the target of image caption optimization is the cross-entropy loss function, which causes the exposure bias problem. The exposure bias problem is that the words generated in the previous step are used during the testing, which is different from the direct use of ground truth during training. If the current prediction is wrong, it will affect all subsequent generation results. In addition, the model uses cross-entropy as the loss function during training, which is inconsistent with the final metrics such as CIDEr\cite{vedantam2015cider}. Therefore, in order to solve the above problems, researchers proposed the method of using SCST\cite{rennie2017self} reinforcement learning. For an image, the SCST algorithm generates two captions, one as the result of greedy decoding and the other as the result of random sampling. CIDEr is used to calculate the reward values of the two, and the advantage is defined as the difference between the reward of the sampled sentence and the reward of the greedy sentence. If the reward of the sampled sentence is higher than that of the greedy sentence, its generation probability is increased. Otherwise, it is reduced. Through this method, not only the exposure bias problem is solved, but also the metrics is directly used as the optimization target to ensure its consistency.

However, the SCST algorithm currently has certain limitations. SCST only relies on a single greedy decoding result as a baseline. If the model itself is not stable enough, the greedy decoding result may be relatively poor, which will lead to a high variance in the advantage estimate, further leading to unstable policy updates. Moreover, SCST only compares one sampling result with the greedy decoding result, resulting in limited diversity of generation and may fall into a local optimum. In addition, when optimizing the model, SCST does not set a reference model and use KL divergence for constraints, which may also make the model more prone to collapse.

Considering the above problems of SCST, we proposed to use the group relative plicy optimization (GRPO)\cite{shao2024deepseekmath} reinforcement learning algorithm to optimize the image captioning model. GRPO is a reinforcement learning algorithm recently proposed by the DeepSeek team, which is designed for reinforcement learning fine-tuning scenarios of large language models. GRPO avoids relying on a separate value network by generating multiple candidate outputs for each input prompt and calculating the advantage function based on the relative reward difference of the output within the group. The core idea of GRPO is to generate multiple candidate answers for the input prompt, and then continuously optimize the model through intra-group comparison, and by constraining the amplitude and KL divergence of the strategy update, the stability of the model during the training process is greatly guaranteed. Therefore, the GRPO algorithm is very suitable for the field of image captioning. Even if the overall reward is sparse, the model can still be optimized through intra-group differences. For example, in a group that generates more incorrect captions, the relative advantage can effectively distinguish between partially correct and completely wrong outputs, and maximize the guarantee that the model is optimized in the right direction. In addition, compared to SCST which only samples one answer, GRPO samples multiple answers. Multiple candidates within a group generate a wider solution space. Combined with the KL divergence constraint, GRPO can improve diversity while ensuring fluency.

We validate our model on MSCOCO2014\cite{lin2014microsoft} offline "Karpathy"\cite{karpathy2015deep} test split. The results show that GRPO reinforcement learning algorithm achieves very good results. The contributions of our paper are summarized as follows:
\begin{itemize}
\item We propose to use the GRPO reinforcement learning algorithm to optimize the image captioning model. Compared with the SCST algorithm, GRPO can sample multiple answers and use KL divergence to constrain the model, which improves the diversity of the model while ensuring accuracy and stability.
\item We conducted experiments on MSCOCO2014. The results show that compared with the SCST method, our method has achieved significant improvements in the image captioning metrics, proving the superiority of GRPO.
\end{itemize}

\section{Method}
\subsection{Image Captioning Model}
\begin{figure}[ht]
\centerline{\includegraphics[width=0.8\linewidth]{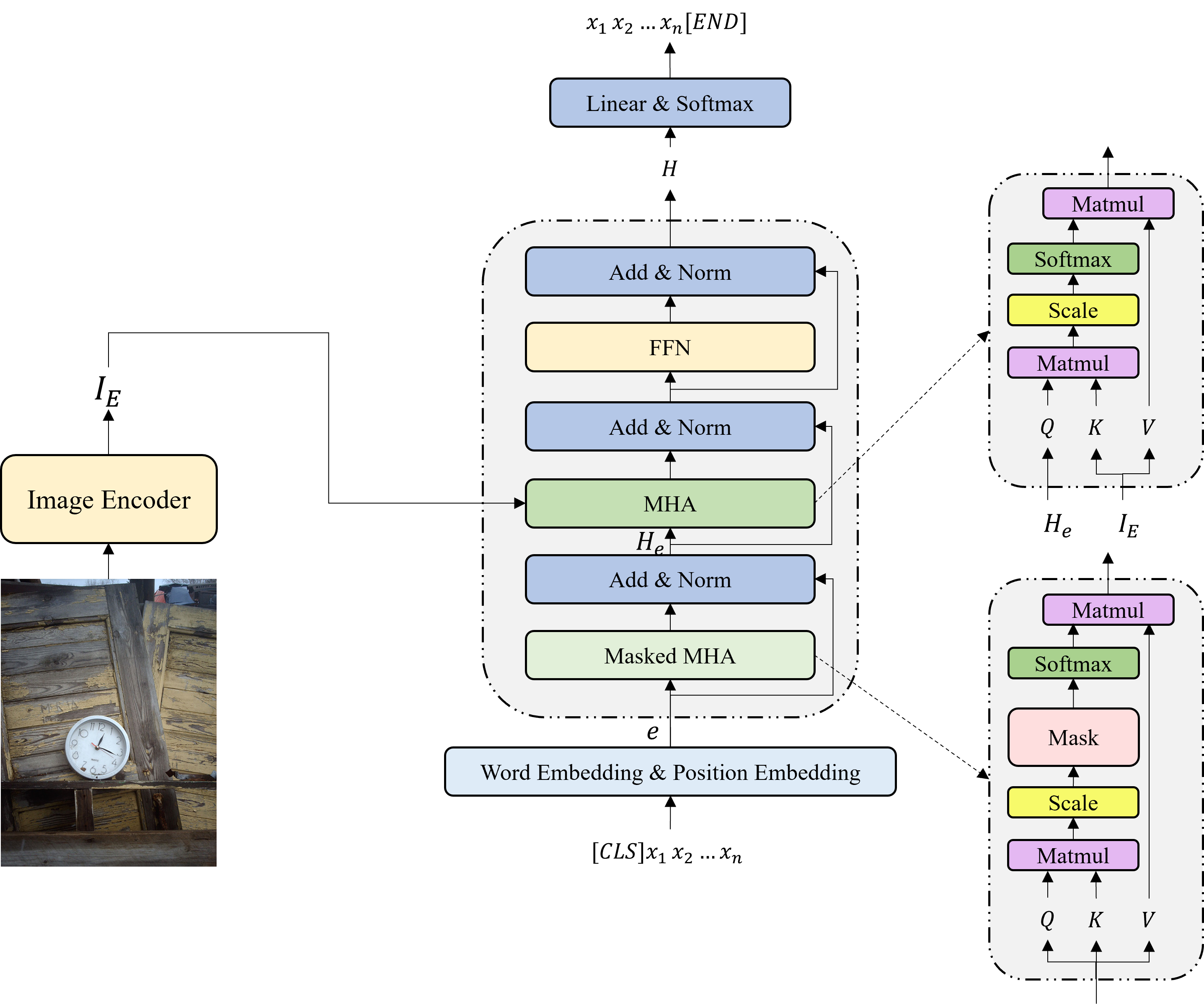}}
\caption{The overall architecture of our image captioning model.}
\label{fig:1}
\end{figure}

The overall architecture of our model is shown in Figure \ref{fig:1}. Consistent with previous work, we still use the encoder-decoder architecture and use the autoregressive method to generate corresponding captions for images. The encoder is a pre-trained CNN network. In this paper, we use the pre-trained ResNet50 as the encoder, and the decoder is the Transformer Decoder. The image features are input into the decoder to complete the decoding and generation of captions in an autoregressive manner.
We first input the image into the image encoder to obtain its feature. The image encoder is a pre-trained ResNet50. The calculation process is as follows:
\begin{equation}
    I_e = CNN\left(C\right)
\end{equation}
Where $I$ is the original image, $CNN$ is the pre-trained ResNet50, and $I_e$ is the extracted image features.

After obtaining the image features, we can start decoding the text. When decoding the text, first add $[CLR]$ as the starting symbol to the input text, generate a position number, and then input it into the word embedding module. The word embedding module consists of a word embedding network and a position embedding network. Since Transformer operates based on the self-attention mechanism and does not contain the position information between texts, it is necessary to add additional position encoding to provide the position relationship. We use the same position encoding method as BERT\cite{devlin2018bert}. Following the word embedding module, we get the text feature and input it into the text decoder together with the image feature. The calculation process is as follows:
\begin{equation}
    e_w = Embedding\left(C\right)
\end{equation}
\begin{equation}
    e_p = Position\_Embedding\left(P\right)
\end{equation}
\begin{equation}
    e = LayerNorm(e_w + e_p)
\end{equation}
Where $e_w$ is the text feature obtained by the word embedding network, $e_p$ is the position information, and $e$ is the text feature after adding the position relationship.

Before fusing text features with image features, we need to process the input text using a masked multi-head attention network. A special attention matrix is introduced in the masked multi-head attention network to ensure that the model can only generate current predictions from past observations by blocking out future information. It is more consistent with the assumption of causality. The attention matrix $M$ generation method is as follows: For any two positional words $i$ and $j$, where $i$ and $j$ represent different position numbers, if $i$ < $j$, then $m_{ij} = 0$, otherwise $m_{ij} = -\infty$. The defined formula is shown below
\begin{equation}
    M = \left\{
    \begin{array}{ll}
        m_{ij} = 0, & i < j  \\
        m_{ij} = -\infty, & otherwise \\
    \end{array} \right.
\end{equation}
Then we use the masked multi-head attention network to process the text features. The calculation formula is as follows:
\begin{equation}
    Masked\_MHA\left(e,M\right) = Concat\left({head}_1,\cdots,{head}_n\right)W^M
\end{equation}
\begin{equation}
    {head}_i\left(e,M\right) = SA\left(e,e,e,M\right)
\end{equation}
\begin{equation}
    SA\left(Q,K,V,M\right) = softmax\left(\frac{QK^T}{\sqrt{d_k}}+M\right)V
\end{equation}
\begin{equation}
    \begin{array}{rl}
        H_e & = Add\&Norm\left(e,Masked\_MHA\left(e,M\right)\right) \\
            & = LayerNorm(e+Masked\_MHA\left(e,M\right))
    \end{array}
\end{equation}
Where $e$ is the word feature obtained after the word embedding module, $W^M$ is the parameters of the network, and $H_e$ is the final text feature obtained.

Finally, we fuse the image features with the text features and input them into FFN to get the final output result. The calculation process is as follows:
\begin{equation}
    MHA\left(H_e,I_e\right) = Concat\left({head}_1,\cdots,{head}_n\right)W^G
\end{equation}
\begin{equation}
    {head}_i\left(H_e,I_e\right)=softmax\left(\frac{H_eI_e^T}{\sqrt{d_k}}\right)I_e
\end{equation}
\begin{equation}
    \begin{array}{rl}
        H^\prime & = Add\&Norm\left(H_e, MHA\left(H_e,I_e\right)\right) \\
                 & = LayerNorm\left(H_e + MHA\left(H_e,I_e\right)\right)
    \end{array}
\end{equation}
\begin{equation}
    H = Add\&Norm\left(H^\prime, FFN\left(H^\prime\right)\right)
\end{equation}
Where $W^G$ is the parameters of the network, and $H$ is the result of fusion feature.

Finally, the fusion features are input into the text predictor to obtain the final prediction result. The text predictor consists of a fully connected layer and Softmax, and generates captions through an autoregressive method.

\subsection{Self-critical Sequence Training}
SCST addresses two fundamental limitations of cross-entropy training in image captioning: (1) exposure bias caused by the discrepancy between teacher-forcing during training and autoregressive decoding at inference, and (2) metric misalignment where cross-entropy optimization fails to directly improve task-specific metrics like CIDEr. By reformulating caption generation as a policy gradient problem, SCST directly optimizes sequence-level evaluation metrics through reinforcement learning. For the input image $I$, the SCST generates two captions using greedy decoding and random sampling. Greedy decoding selects the highest-probability token at each timestep, while random sampling draws tokens from the predicted probability distribution. The central idea of the self-critical sequence training approach is to baseline the REINFORCE algorithm with the reward obtained by the current model under the inference algorithm used at test time. The calculation method is as follows:
\begin{equation}
    L_R\left(\theta\right)=-E_{x_{1:T} \sim p_\theta}\left[r\left(x_{1:T}\right)\right]
\end{equation}
Where $r\left(\cdot\right)$ is the score of CIDEr, $\theta$ is the model parameter. The gradient of $L_R$ can be approximated as follows:
\begin{equation}
    \nabla_\theta L_R\left(\theta\right)\approx-\left(r\left(x_{1:T}^s\right)-r\left({\hat{x}}_{1:T}\right)\right)\nabla_\theta\log{p_\theta\left(x_{1:T}^s\right)}
\end{equation}
Where $x_{1:T}^s$ is a sampled caption and ${\hat{x}}_{1:T}$ is a the greedily decoded caption. This self-baseline approach eliminates the need for separate reward models while ensuring direct optimization of evaluation metrics. While SCST effectively bridges the gap between cross-entropy training and metric-driven optimization, its design also has certain limitations in image captioning. SCST does not use KL divergence as a constraint, so it is prone to mode collapse. And it only generates one sampling result, so the diversity may also be poor.

\subsection{Group Relative Policy Optimization}
GRPO reinforcement learning is a reinforcement learning method proposed by the DeepSeek team. It eliminates the need for an additional value function like in PPO, and instead uses the average reward value of multiple sampled outputs as its baseline for the same input. More specifically, for each question $q$, GRPO extracts a set of outputs $\{o_1, o_2, \ldots, o_G\}$ from the old strategy $\pi_{\theta_{old}}$, and then optimizes the policy model by maximizing the following objectives.
\begin{equation}
    \begin{array}{ll}
        \mathcal{J}_{GRPO}\left(\theta\right) & = \mathbb{E}\left[q \sim P\left(Q\right), \{o_i\}_{i = 1}^G \sim \pi_{\theta_{old}}\right] \\
        & \frac{1}{G} \sum_{i=1}^n \left( \min (\frac{\pi_\theta(o_{i}|q)}{\pi_{\theta_{\text{old}}}(o_{i}|q)} A_i, \text{clip}\left( \frac{\pi_\theta(o_{i}|q)}{\pi_{\theta_{\text{old}}}(o_{i}|q)}, 1-\epsilon, 1+\epsilon \right) A_i ) - \beta \mathbb{D}_{\text{KL}} \left( \pi_\theta \big\| \pi_{\text{ref}} \right) \right)
    \end{array}
    \label{formula:16}
\end{equation}
\begin{equation}
    \mathbb{D}_{\text{KL}} \left( \pi_\theta \big\| \pi_{\text{ref}} \right) = \frac{\pi_{\theta_{\text{ref}}}(o_{i}|q)}{\pi_\theta(o_{i}|q)} - \log \frac{\pi_{\theta_{\text{ref}}}(o_{i}|q)}{\pi_\theta(o_{i}|q)} - 1
\end{equation}
where $\epsilon$ and $\beta$ are hyper-parameters, and $A_i$ is the advantage, computed using a group of rewards ${r_1, r_2, \ldots, r_G}$ corresponding to the outputs within each group:
\begin{equation}
    A_i = \frac{r_i - mean(\{r_1, r_2, \ldots, r_G\})}{std(\{r_1, r_2, \ldots, r_G\})}
\end{equation}
It can be seen that compared with the SCST algorithm, the GRPO algorithm limits the amplitude of the policy update and adds KL divergence as a constraint, which greatly enhances the stability of the model. In addition, GRPO uses multiple sampling results, which also improves the diversity of the generated results. Therefore, we apply the GRPO algorithm to the image captioning task. For the input image $I$, we also sample and generate multiple answers, and use CIDEr as its reward value, and then optimize the model according to Formula \ref{formula:16}.

\section{Experiments}
\subsection{Dataset and Evaluation Metrics}
We use MSCOCO2014 as our experiment dataset. MSCOCO2014 contains a rich variety of images and their associated captions. It contains 123,287 images, of which 82,783 images are classified into the training set and 40,504 are classified into the validation set. There are 5 reference captions for each image, which cover a wide range of scenarios and topics, including characters, animals, natural landscapes, indoor environments, etc. In this paper, we follow the “Karpathy” split to redivide the MSCOCO, where 113287 images for training, 5000 images for validation and 5000 images for evaluation. 

In order to evaluate the captions quality generated by the model, we use five common metrics, including BLEU\cite{papineni2002bleu}, METEOR\cite{agarwal2008meteor}, ROUGE-L\cite{lin2004rouge}, CIDEr\cite{vedantam2015cider} and SPICE\cite{anderson2016spice}. The following are the calculation formulas for these metrics.
\begin{equation}
\text{BLEU} = \text{BP} \cdot \exp\left( \sum_{n=1}^{N} w_n \log p_n \right)
\end{equation}
\begin{equation}
\text{METEOR} = F_{\text{mean}} \cdot (1 - \text{penalty})
\end{equation}
\begin{equation}
\text{ROUGE-N} = \frac{\sum_{\text{ngram} \in \text{ref}} \min(\text{count}_{\text{gen}}(\text{ngram}), \text{count}_{\text{ref}}(\text{ngram}))}{\sum_{\text{ngram} \in \text{ref}} \text{count}_{\text{ref}}(\text{ngram})}
\end{equation}
\begin{equation}
\text{CIDEr} = \frac{1}{m} \sum_{i=1}^{m} \frac{\sum_{j=1}^{n} \text{TF-IDF}_i(g_j) \cdot \text{TF-IDF}_i(r_j)}{\sqrt{\sum_{j=1}^{n} \left( \text{TF-IDF}_i(g_j) \right)^2} \cdot \sqrt{\sum_{j=1}^{n} \left( \text{TF-IDF}_i(r_j) \right)^2}}
\end{equation}
\begin{equation}
\text{SPICE} = \frac{1}{K} \sum_{k=1}^{K} F_{\text{score}}(S_k, \hat{S}_k)
\end{equation}
Where BP is a brevity penalty term used to penalize excessively short generated texts, $p_n$ is The precision of the nth n-gram between the generated text and the reference text, $w_n$ represents the weight, which is often equal to $1/N$, $N$ is the maximum length of the n-gram, $F_{\text{mean}}$ is a harmonic mean of precision and recall, $\text{penalty}$ is a penalty term for mismatched sequences. $g_j$ represent the n-grams of the generated captions, $r_j$ represent the n-grams of the reference captions, $\text{TF-IDF}_i$ is the TF-IDF value of the i-th word in the generated and reference captions, $S_k$ and $\hat{S}_k$ are the semantic parsing graphs of the reference caption and the generated caption, respectively, and $F_{\text{score}}$ is the F1 score for semantic matching.

\subsection{Experimental Settings}
In this paper, we use the pre-trained ResNet50 as the image encoder, the image size is uniformly resized to $224\times 224$, the word embedding size is set to 512, and the number of Transformer decoder layers is 1. The word dictionary of BERT is used in this experiment. During the training process, cross-entropy is first used as the objective function for 20 epochs. The learning rate is 4e-5. The batch size is set to 32, the warm-up step is $1/10$ of the total number of iterations, and then cosine decay is used. Then the model weights saved in the last epoch are used to initialize the starting model of the SCST and GRPO algorithms, and reinforcement learning strategy optimization is performed on this basis. The learning rate of SCST and GRPO algorithms is set to 1e-5, and cosine decay is used, and the batch size is 32. SCST is trained for 20 epochs. GRPO is trained for 5 epochs, the update step is set to 20, and the number of sampled answers is set to 5. In the validation and evaluation process, the beam search decoding method is used with the beam size set to 3. Our experiments used the mindspore framework and implemented the SCST and GRPO algorithms for the image captioning.

\subsection{Result Analysis}
As shown in Table \ref{tab:1}, we give the results of cross-entropy optimization and the optimization results of SCST and GRPO. It can be seen that GRPO exceeds SCST in all metrics, with BLEU-4 being 0.9\% higher and CIDEr being 2.4\% higher, which proves the advantage of the GRPO algorithm. It can be seen that the GRPO algorithm significantly improves CIDEr, which is consistent with our use of CIDEr scores as rewards. In addition, from the experimental setting, we only trained the GRPO algorithm for 5 epochs and achieved better results than the SCST algorithm, which also shows the efficiency of the GRPO algorithm.

\begin{table}[ht]
\caption{Experiment results based on MindSpore 2.2.14 framework on MSCOCO “Karpathy” test split.}
\centering
\resizebox{0.9\linewidth}{!}
{
\begin{tabular}{lllllllll}
\hline
Optimization Methods & BLEU-1        & BLEU-2        & BLEU-3        & BLEU-4        & METEOR        & ROUGE-L       & CIDEr          & SPICE         \\ \hline
CE                   & 66.9          & 49.5          & 36.3          & 26.8          & 23.7          & 50.5          & 84.3           & 16.4          \\
+SCST                & 74.0          & 56.5          & 41.6          & 30.5          & 23.8          & 52.5          & 97.6           & 16.4          \\
+GRPO                & \textbf{75.0} & \textbf{57.7} & \textbf{42.8} & \textbf{31.4} & \textbf{24.4} & \textbf{53.2} & \textbf{100.0} & \textbf{17.1} \\ \hline
\end{tabular}
}
\label{tab:1}
\end{table}

\begin{table}[ht]
\caption{Experiment results based on MindSpore 2.2.14 framework on Flickr8k test dataset.}
\centering
\resizebox{0.9\linewidth}{!}
{
\begin{tabular}{lllllllll}
\hline
Optimization Methods & BLEU-1        & BLEU-2        & BLEU-3        & BLEU-4        & METEOR        & ROUGE-L       & CIDEr         & SPICE         \\ \hline
CE                   & 55.8          & 36.4          & 23.2          & 14.8          & 17.2          & 41.5          & 34.2          & 10.9          \\
+SCST                & 58.1          & 39.0          & 25.3          & 16.4          & 16.4          & 41.9          & 36.6          & 10.6          \\
+GRPO                & \textbf{60.6} & \textbf{40.4} & \textbf{25.9} & \textbf{16.7} & \textbf{17.5} & \textbf{43.0} & \textbf{38.3} & \textbf{11.1} \\ \hline
\end{tabular}
}
\label{tab:2}
\end{table}

In addition, in order to verify the stability of the SCST and GRPO algorithms, we conducted experiments using the Flickr8k\cite{rashtchian2010collecting} dataset. Flickr8k contains 8,000 images, of which 6,000 are used for training, 1,000 for validation, and 1,000 for testing. Each image also has 5 associated captions. Since the dataset contains only a small number of images, the model capability after cross-entropy optimization may be slightly weaker. The experimental results are shown in Table \ref{tab:2}. It can be seen that GRPO still has an improvement in all metrics, while SCST has a decrease in some metrics. In addition, in the experiment, we also found that the SCST algorithm occasionally crashes, that is, the validation set metrics suddenly drop sharply, while the GRPO algorithm is very stable. Even if the baseline model is slightly weaker, GRPO can still steadily improve the performance of the model, which also verifies its stability.

\section{Conclusion}
In this paper, we propose the GRPO reinforcement learning algorithm for image captioning that addresses the limitations of conventional SCST through group-wise optimization and distributional constraints. By generating multiple candidate captions per image and optimizing their relative quality through intra-group comparisons, GRPO achieves a better exploration-exploitation trade-off. The KL divergence constraint effectively prevents semantic drift, greatly improving the stability of the model. Experimental results on the MSCOCO2014 and Flickr8k datasets show that the GRPO algorithm can stably and efficiently improve the capabilities of the model.

\section*{Acknowledgments}
Thanks for the support provided by MindSpore Community. All experiments proposed in this paper are implemented based on the mindspore framework.

\bibliographystyle{unsrt}  
\bibliography{references}  






\end{document}